\pdfoutput=1 
\documentclass[
]{ceurart}

\usepackage{listings}
\usepackage{subcaption}

\begin{document}

\copyrightyear{2024}
\copyrightclause{Copyright for this paper by its authors.
  Use permitted under Creative Commons License Attribution 4.0
  International (CC BY 4.0).}

\conference{HAII5.0: Embracing Human-Aware AI in Industry 5.0, at ECAI 2024, 19 October 2024, Santiago de Compostela, Spain.}

%%
%% The "title" command: please use "title case capitalisation" 
%% Guidelines can be found here: https://wiki.musicbrainz.org/Style/Language/English
\title{Augmenting train maintenance technicians with automated incident diagnostic suggestions}
%\tnotetext[1]{You can use this document as the template for preparing your publication. We recommend using the latest version of the ceurart style.}

\author[]{Georges Tod}[%,
email=georges.tod@sncb.be,]
\cormark[1]
\address[]{SNCB-NMBS Engineering Department, Brussels, Belgium }

\author[]{Jean Bruggeman}
\author[]{Evert Bevernage}
\author[]{Pieter Moelans}
\author[]{Walter Eeckhout}
\author[]{Jean-Luc Glineur}

\cortext[1]{Corresponding author.}
%\fntext[1]{These authors contributed equally.}

\begin{abstract}
Train operational incidents are so far diagnosed individually and manually by train maintenance technicians. In order to assist maintenance crews in their responsiveness and task prioritization, a learning machine is developed and deployed in production to suggest diagnostics to train technicians on their phones, tablets or laptops as soon as a train incident is declared. A feedback loop allows to take into account the actual diagnose by designated train maintenance experts to refine the learning machine. By formulating the problem as a discrete set classification task, feature engineering methods are proposed to extract physically plausible sets of events from traces generated on-board railway vehicles. The latter feed an original ensemble classifier to class incidents by their potential technical cause. Finally, the resulting model is trained and validated using real operational data and deployed on a cloud platform. Future work will explore how the extracted sets of events can be used to avoid incidents by assisting human experts in the creation predictive maintenance alerts.
\end{abstract}


\maketitle

\section{Introduction}
The last ten years in the rolling stock community has seen a shift from purely on-board diagnostics by human experts to assisted human diagnostics by remote diagnostics. As a matter of fact, recent railway vehicles are equipped with sensors on most of their subsystems such as pantographs, traction converters, doors, heating, ventilation and air-conditioning (HVAC), European Train Control System (ETCS)\footnote{the ETCS helps drivers to operate the trains safely}, etc. which report some states via a wired network to a central on-board computer. The latter typically transmits some of these states or combinations of these states as tokens to a cloud service over a cellular network. The tokenized states are often nominated as \textit{information codes} (e.g. \texttt{door 1 is open}) or \textit{fault codes} (e.g. \texttt{battery temperature is too high}) and represent \textit{events} that happen on-board the vehicles. So far, when an incident happens during train operations, a human expert needs to read the traces of events to interpret what was the cause of the incident. While individual events are informative of incidents, it is usually a set of events and their context that can explain the cause of an incident. A loose analogy with human language can be made. Information and fault codes could be seen as the vocabulary of railway vehicles and the more there are, the more expressive the vehicle language will be. However, this language is far from being trivial and requires time for humans to learn it and being able to diagnose incidents. As a result, diagnosing the causes of incidents scales so far at the pace of human availability and training. When train operators fleets' grow, the number of incidents grows and a bottleneck arises. In the context of railway vehicles maintenance, when an incident happens and a given vehicle needs to reach a workshop, it requires some planning to decouple vehicles. Furthermore, workshops might be specialized in certain maintenance tasks. A preliminary diagnostic available close to real-time will also help prioritizing which vehicles need to reach which workshop and when. If a failure is not safety related, it can also be decided to postpone the immobilization of a vehicle and let it operate until further notice. Better informed decisions about the health of vehicle will help increase the number of vehicles available for operations. To achieve this and support train maintenance technicians in their diagnostics, a learning machine is proposed in this work to: (1) extract meaningful recurrent sets of events and (2) propose a technical cause to an incident based on the latter. To do so, the problem is formulated as a classification task, for which both the feature extraction and the result of the classification are of importance,
\begin{equation}
y = f(x)
\end{equation}
where $x = [x_1,x_2,...,x_p]$ is a set of events that are generated around the timestamp of an incident and $y$ the physical subsystem (ETCS, high or low voltage equipment, doors, etc.) that caused the incident. For example, a set of events could be: \texttt{[train speed is higher than 0, speed check by beacon invalid, requesting automatic emergency braking]}. A label $y$ for the latter, could be related to a failure of the \texttt{European Train Control System} (ETCS). In addition, a human feedback loop composed of train maintenance experts is setup to improve both the training data quality $x,y$ and the function $f$. Aside from the classification, train maintenance experts have requested the need to understand scenarios that lead to an incident in order to create remote diagnostics alerts using deterministic rules. Future work will evaluate if the extracted events in $x$ can help designing such alerts.

\paragraph{Related works} In \cite{thelen2022comprehensive} a review of how digital twins can be used in the context of diagnostics and fault identification is proposed. The field is quite advanced when vibration data can be leveraged, see \cite{randall2021vibration}. Deep learning techniques are more and more used to monitor and diagnose the health of machines  without the need of feature engineering methods, see \cite{zhao2019deep}. Applications of machine learning to railways systems can be found in \cite{ma2019deep}, where railway track quality is assessed using deep learning. In \cite{fink2013predicting}, restricted Boltzmann machines and echo state networks are combined to predict the occurrence of railway operation disruptions. In \cite{dalle2022machine}, railway planning problems are treated using machine learning. More fundamentally, when it comes to the analysis of discrete sequences outlier detection methods are reviewed in \cite{chandola2010anomaly} and \cite{aggarwal2016outlier}. Hidden Markov models (HMMs) have been successfully applied to detect intrusions based on sequential data in \cite{gao2002hmms}. Long short-term memory networks (LSTMs) are applied in \cite{zhou2015c} and \cite{karim2019multivariate} for text and timeseries classification. More recently, Deterministic Finite Automata (DFA) seem to achieve similar performance with better interpretability, see \cite{shvo2021interpretable}. Fault identification based on timeseries is a more mature field than based on discrete sequences of events with nonuniform sampling. Furthermore, it is a necessity for safety reasons not only to identify a fault but also to be able to explain why it is qualified as such.\\

To the best of our knowledge, our problem has not been yet formulated in literature. Therefore, the following contributions are claimed: (1) the formulation of automated diagnostics of railway vehicle incidents based on on-board generated events as a classification task, (2) feature engineering methods to extract recurrent sets of events from traces generated on-board railway vehicles and (3) a novel discrete set classification algorithm is proposed to solve our problem.

\section{Methodology}
The suggestion of incident diagnostics problem is casted as a supervised learning problem deployed on a cloud platform. In the next section, the data sources for labels and features, the platform used for deployment and how the learning machine interfaces with humans within in place maintenance processes are described. In the second section, the learning machine design is detailed.

\subsection{An automated diagnostics platform}
\begin{figure}
\centering
	\includegraphics[clip,trim={0cm 10cm 0cm 0cm},width=15cm]{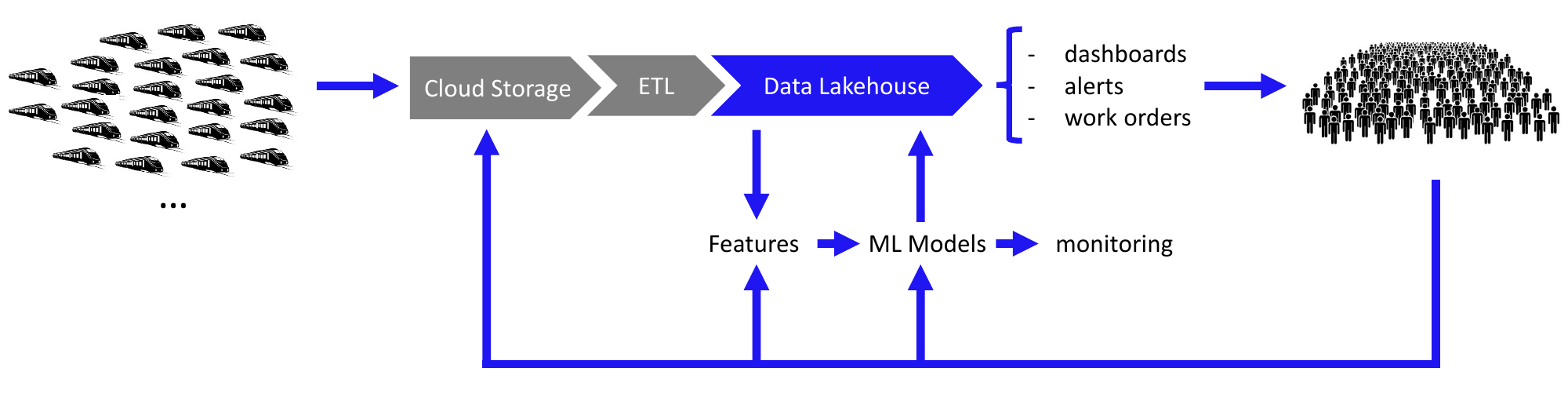}
\caption{\textbf{Platform to assist train maintenance technicians} by automatically suggesting incident diagnostics of railway vehicles. Railway vehicles central on-board computers report about their states to a cloud storage. By Extracting-Transforming and Loading (ETL) the raw data, structured data is fed into a Data Lakehouse. Iterable machine learning models leverage iterable features to analyze large volumes of data and deliver online dashboards to assist train maintenance technicians. A loop allows to take into account the feedback from designated train maintenance experts diagnostics to refine both the training data and the models.}
\label{fig:platform}
\end{figure}

\paragraph{Data sources and characteristics} In order to train the proposed classifier, railway vehicles incidents and events need to be mined.
The incidents considered happen during train operations and are severe enough to provoke a train delay of at least 6 minutes. Mitigating these is of major importance for train operators. Incident datasets are imbalanced as there is no reason for all types of technical issues to happen as often. When an incident is declared, a cloud platform (figure \ref{fig:platform}) ingests both its timestamp and train composition. The latter are used to extract the sequences of events around the timestamps for the vehicles in the train composition (figure \ref{fig:feats}) as detailed in the next section.\\
The raw events that will be used to build the features $X$ are generated on-board railway vehicles. An event represents a behavior which is dependent on the software deployed on-board. Some railway vehicles, typically locomotives report additional detailed events about their traction systems that passenger vehicles do not report; the different kinds of events can be seen as different languages expressed by vehicles and all of them need to be processed by the learning machine.\\
The classification task involves the 12 following labels  $y$ or physical subsystems : \textit{ETCS, high or low voltage equipment, couplings, doors, brakes, communication, air production, cabling, body, traction, sanitaries or others}. Samples are labeled by train maintenance technicians after in-depth analysis when vehicles reach the workshops. The cloud platform (figure \ref{fig:platform}) also ingests these analysis. Since the data is logged by a large diverse crew, its biases are multiple, adding a supplementary layer of complexity.

\paragraph{Cloud platform}
An expressive fleet of 300 vehicles generates around 300$k$ events per day. For such a fleet, the number of monthly incidents ranges between 50 and 100. In order to process such data flow for different fleets, the platform described on figure \ref{fig:platform} is developed and maintained. \\
During their lifetime, railway vehicles need their on-board software to be updated for operational and safety reasons. In addition, on the long run, some new technical issues will appear and well treated ones, will disappear. As a result, the learning machine will need to be re-trained during its life cycle to adapt to both time-varying vehicle behavior and incident types. To do so, machine learning operations principles (MLOps) \cite{sculley2015hidden}, such as asynchronous updates of features or models are taken into account, see figure \ref{fig:platform}\footnote{in practice we use the \textit{mlfow} library in python}.\\
The output of the proposed learning machine pushes its suggestions to online dashboards that can be consulted from any tablet, smartphone or laptop within the company. Any train technician can therefore consult suggestions at any time. 

\paragraph{Feedback loop} Human experts are designated among the maintenance crews to identify a single source of truth whenever the learning machine and train maintenance technicians disagree. In a later stage, the learning machine can be re-trained (MLOps) using the resulting larger and larger high quality dataset for which the cause of the incident is certain. The same experts help refining the feature engineering methods and the design of algorithms. Typical refinements from experts allow to filter on the context of events before these are taken into account.\\

It must be noted, that the learning machine is capable of producing a diagnostic remotely as soon as an incident has been declared and the events related to it have been ingested by the proposed platform. The dynamics of this process are therefore much faster than the dynamics of the train maintenance technicians diagnostics process reporting: seconds versus days. The next sections discuss how the proposed learning machine is designed and performs.

\subsection{A discrete set classification algorithm}

In this section, first the two stages of the proposed feature engineering methods are explained: (1) the filtering of raw events and (2) the extraction of event sets. Second an original ensemble classifier based on the extracted sets is proposed.

\subsubsection{Feature engineering}

\begin{figure}
 \centering
	\includegraphics[clip,trim={1cm 12cm 2.5cm 0cm},width=15cm]{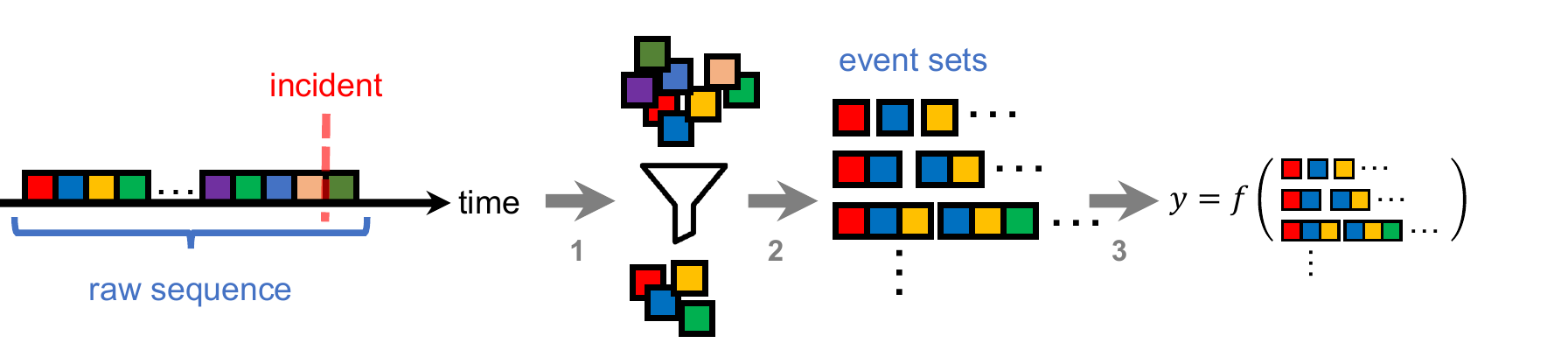}
\caption{\textbf{Feature engineering}: (1) events are filtered based on a \textit{relevance} metric $r$ and a \textit{One-at-a-time (OaT)} procedure. In (2), events sets are mined based on \textit{Longest-Common Sub Sequences (LCSS)}. The latter are fed in (3) to the proposed ensemble classifier.}
\label{fig:feats}
\end{figure}

\begin{figure}
 \centering
\begin{subfigure}[b]{0.3\textwidth}
         \centering
	\includegraphics[width=4.5cm]{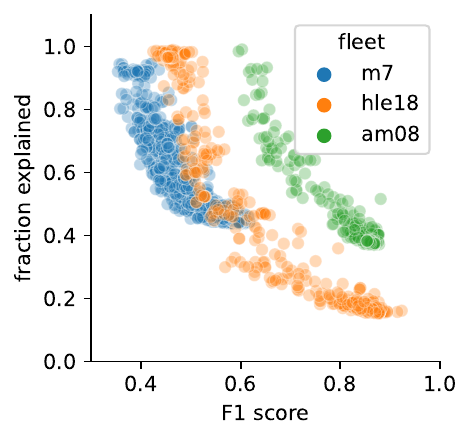}
         \caption{}
\end{subfigure}
\begin{subfigure}[b]{0.6\textwidth}
         \centering
	    \includegraphics[width=9cm]{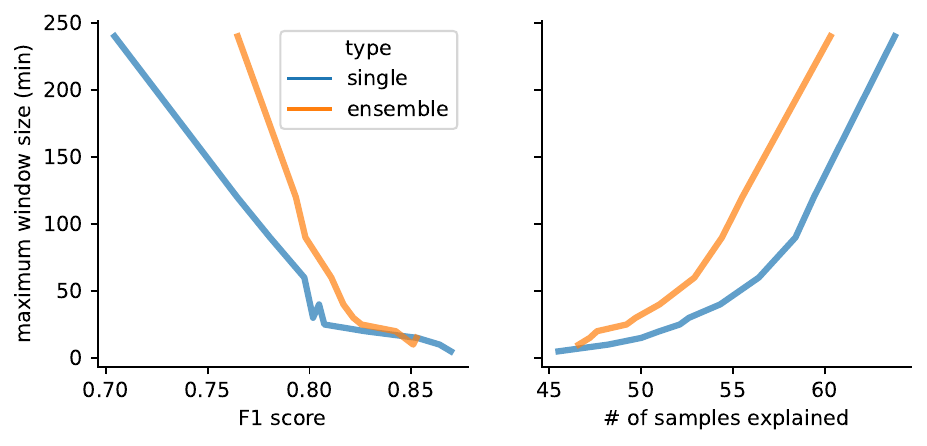}
         \caption{}
\end{subfigure}
\caption{\textbf{Hyperparameter exploration results}. In (a) individual features performances for the \textit{One-at-a-time (OaT)} procedure. The fraction of explained samples is the mean ratio of number of classified samples over the total number of samples on the 10 folds of the stratified cross validation. In (b), the performance of single versus the proposed ensemble classifier. The number of explained samples is the mean number of classified samples on the 10 folds of the stratified cross validation. Interestingly, the larger the window, the more the $F_1$-score drops: meaning the further in the past the model looks at, the less it can leverage the additional data. Any event happening earlier than four hours before an incident is not taken into account as classifiers' performance is considered too low in terms of $F_1$-score.}
\label{fig:tuning}
\end{figure}

\paragraph{Filtering features} In a first stage, events are filtered as not all of them are useful for incident diagnostics, see figure \ref{fig:feats} (a). A \textit{relevance} metric $r$ is used to determine the most informative individual events in a multiclass problem, 
\begin{equation}
\textit{r} = \frac{h_\text{in class}}{h_\text{in all classes}} 
\end{equation}
where $h$ is an event frequency in class or in all classes. The higher $\textit{r}$, the more discriminative across classes an event is. A threshold $t_r$ is tuned\footnote{by a stratified 10 fold cross validation} for $r$ such that the chosen features lead to an $F_1$-score higher or equal to 90\%. Training a classifier with such features gives a classifier with a high $F_1$-score but since many features are discarded, input data can become inexistent resulting in the classification of very few incidents. The \textit{number of classifications} is introduced in addition to the $F_1$-score as a performance metric. It allows to evaluate which additional features can help increasing the number of classified incidents without impacting too much the $F_1$-score.\\
The events that are not retained due to $r<t_r$, go through a second stage denominated \textit{One-at-a-time (OaT)} procedure. It consists in training a classifier for one additional event at a time and evaluating the $F_1$-score and the number of classified incidents. On figure \ref{fig:tuning} (a), the results show that a trade-off between the two needs to be found. It is chosen here to retain all events that maintain the classifier's $F_1$-score higher than 85\%. Interestingly, no additional features are found to be useful for the M7 fleet for which the on-board software is not yet mature and very heterogeneous across the fleet.

\paragraph{Extracting sets of features} Once the events that can constitute a set have been determined, in a second stage, recurring sets of different lengths are mined. In order to do so an estimator of the \textit{Longest-Common Sub Sequence (LCSS)} algorithm is used, see \cite{tslearn2020}. Sensors can sometimes report the same event hundreds of times in a couple of minutes: to denoise the sequences from this effect, sets of events are taken from the raw sequences. For a given maximum sequence\footnote{the sets of events can also be seen as sequences} length, the algorithm scans for the LCSS in historical data. If some subsequence is found, then it is retained as a new feature in addition to the initial one. A very interesting example of the outcome is these three events set: \texttt{[train speed is higher than 0, speed check by beacon invalid, requesting automatic emergency braking]}. The latter is a physically plausible scenario that will lead to an incident and that was automatically extracted using the proposed approach.\\

Based on the features mined using the approaches presented, the design of an original classifier is detailed in the next section.

\subsubsection{Classification}
\begin{figure}
	\centering
	\includegraphics[clip,trim={0cm 5cm 8cm 0cm},width=10cm]{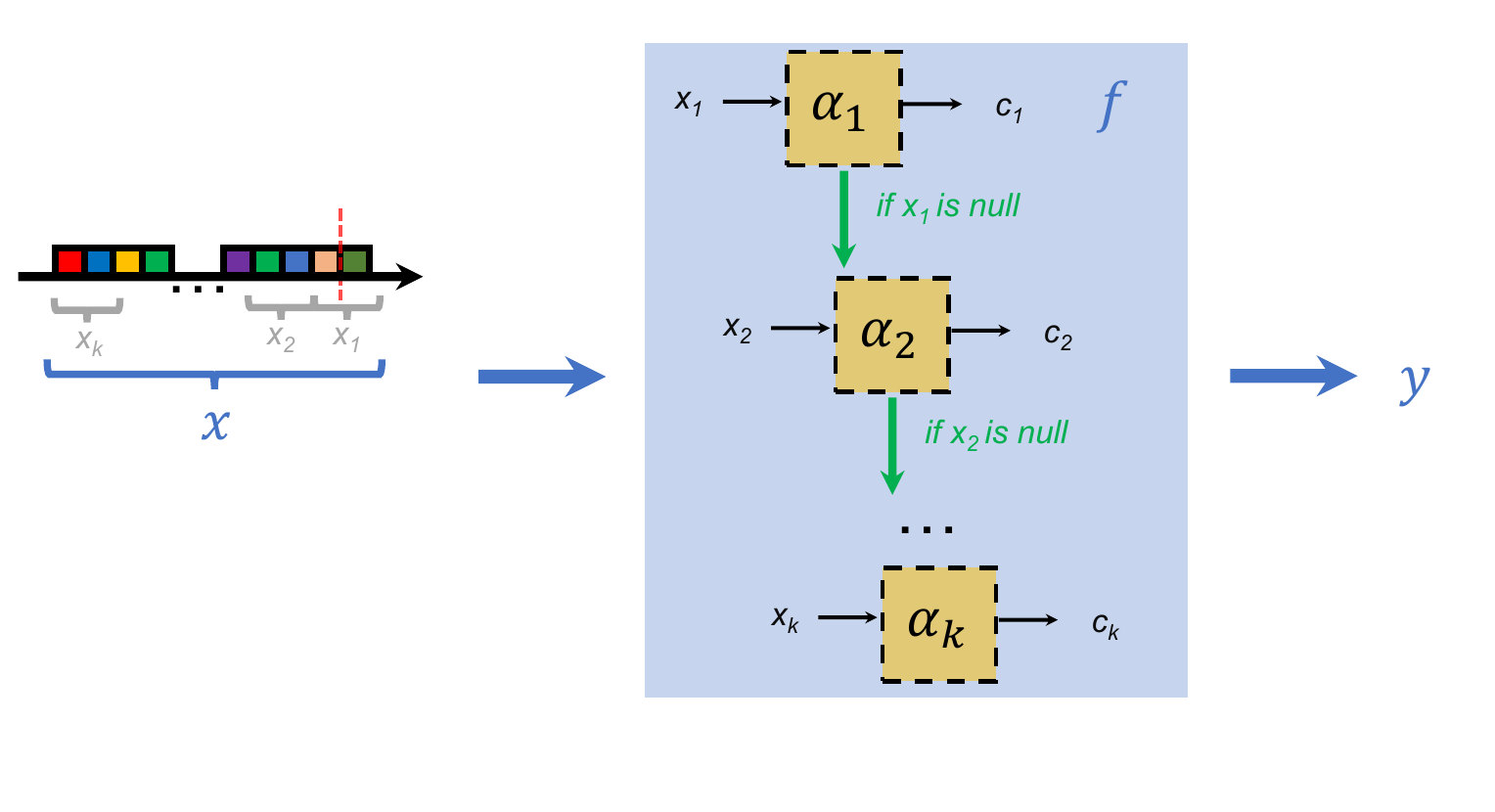}
	\caption{\textbf{Ensemble classifier architecture}: the proposal is based on cascaded time windows. The first classifier to answer fixes the output, which means the collective decision process assumes the first classifier to answer is the most performant one.}
\label{fig:ensembleclass}
\end{figure}

In this section, an original ensemble classifier based on a naïve Bayes classifier is proposed. Each individual classifier $c_k$ estimates the class according to $c_k = \alpha_{k} (x_k)$ where,
\begin{equation}
 \alpha_{k} = argmax_{j} \Big ( p(c_j)\prod_{i=1}^{n_{x_k}} p(x_{k_i}|c_j) \Big )
\end{equation}
and where $x_{k_i}$ are sets of events in the window $x_{k}$ and $n_{x_k}$ is the number of features on that window, see figure \ref{fig:ensembleclass}. The event sets are discrete variables for which the probability of feature $x_{k_i}$ in window $x_k$ given the class $c_j$ is,
\begin{equation}
p(x_{k_i}|c_j) =  \frac{\mathbf{card}(x_{k_i}|c_j)+\beta}{n_{x_k}\beta +\sum_i \mathbf{card}(x_{k_i}|c_j)}
\end{equation}
where $\beta$ is a (small) smoothing parameter to avoid zero probabilities. The design of the ensemble is based on empirical evidence showing that the further the events are from the incident, the higher the error rate of the classifier is. This is illustrated on figure \ref{fig:tuning}(b): single (non-ensemble) classifiers are trained on windows of varying length (see table \ref{tbl:w_sizes}) and performance results show the larger the window, the lower the $F_1$-score. This observation triggered the idea of cascading classifiers over a range of windows according to their position w.r.t. the incident time, see figure \ref{fig:ensembleclass}. It implies the assumption that the first classifier that answers is the most performant. Figure \ref{fig:tuning}(b) shows for an equivalent window length the $F_1$-score superiority of the proposed ensemble classifier with respect to a single one. The number of classified incidents is lower with the ensemble, but it is marginal w.r.t. $F_1$-score improvement. The number of windows $x_k$ (equal to the number of classifiers $k$) and their sizes are tuned by a stratified 10 fold cross validation over a grid (see table \ref{tbl:w_sizes}). The fact that the number of windows and their sizes matter, show the positions of the features with respect to time are important. \\

\begin{table}
\caption{Explored window sizes for both single and the proposed ensemble classifier.}
\label{tbl:w_sizes}
\begin{center}
  \begin{tabular}{ l | l | c }
    \hline
single    & ensemble & max. window size (min)  \\ \hline
    \texttt{[5]}    & -                                                                         & 5  \\ \hline
    \texttt{[10]}  & \texttt{[5,10]}                                                       & 10\\ \hline
    \texttt{[15]}  & \texttt{[5,10,15]}                                                   & 15\\ \hline
    \texttt{[20]}  & \texttt{[5,10,15,20]}                                               & 20\\ \hline
    \texttt{...}     & \texttt{...}                                                            & \texttt{...}\\ \hline
    \texttt{[240]} & \texttt{[5, 10, 15, 20, 25, 30, 40, 60, 90, 120, 240]}    & 240 \\
    \hline
  \end{tabular}
\end{center}
\end{table}

\begin{figure}
	\centering
	\includegraphics[width=15cm]{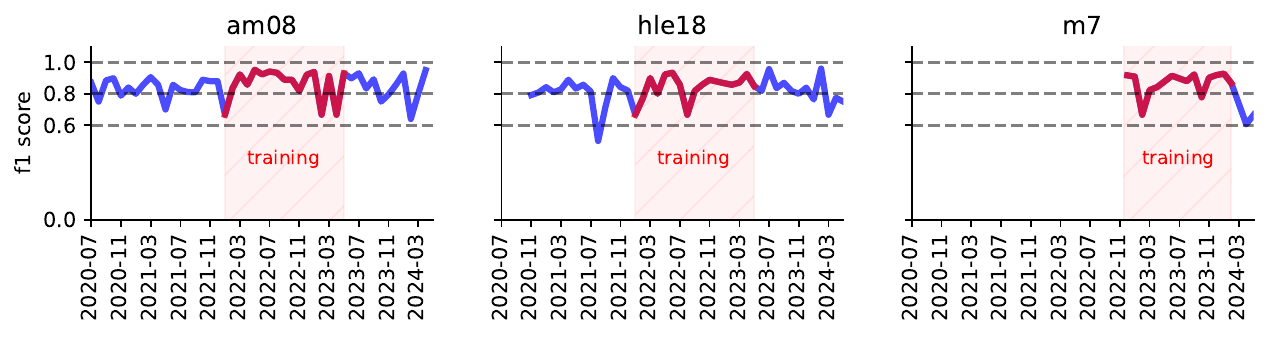}
	\caption{\textbf{Learning machine performance}: descriptive (red) and predictive (blue) performances across three different fleets (AM08, HLE18 and M7): typically the $F_1$-score is high. Nevertheless some incidents are poorly classified even during training. The M7 is a very recent fleet which explains why there is less data.}
\label{fig:pred}
\end{figure}

\section{Results and discussion}
\begin{figure}
	\centering
	\includegraphics[width=9cm]{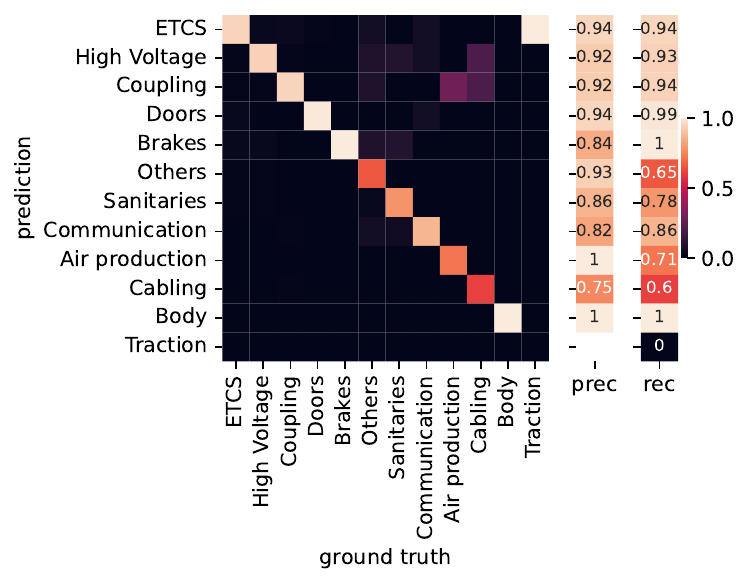}
	\caption{\textbf{Confusion matrix for descriptive performance} of the AM08 fleet: both precision and recall are reported. Subsystems are ordered by their order of importance. Not all subsystems can be classified with the same performance. For this fleet, traction subsystems incidents' do not happen enough to have a sufficiently large dataset (~1 incident per year) leading to misclassifications.}
\label{fig:confmat}
\end{figure}

The models are trained using up to 1.5 years of data from the historical Belgian train operator\footnote{SNCB-NMBS} nationwide operations. For the AM08 fleet, it represents around 140 million raw events and 900 incidents. The models predictive performances are verified for up to 2 years, depending on the available data for a given fleet of vehicles, see figure \ref{fig:pred}. The choice of the training data was motivated by two factors: (1) staying away from Covid-19 low intensity period of operations and (2) choosing a period of low variations of vehicle on-board software versions. In terms of computational cost, the resulting model requires little power\footnote{CPU used : Intel Xeon $2.4$Ghz with $8$ physical cores. It takes $10$s to train our tuned classifier on $1.5$ years of data which we estimate to consume $125$mWh. The prediction of a sample takes around $100$ms which can be estimated as $1.25$mWh.} and runs fast enough to open the potential for edge computing in the future.\\
By computing the $F_1$-score across all classes on out of training datasets, the predictive performance is typically around 80\%, see figure \ref{fig:pred}. It is also clear that some incidents are poorly classified by the models even during training. On figure \ref{fig:confmat} the confusion matrix on the training data for the AM08 fleet gives an idea of why. The performance of the classifier on the most common incidents is high: above $90 \% $ for both precision and recall for European Train Control System (ETCS), high tension equipment, couplings and doors. However, cabling and traction incidents are harder or even impossible to classify using our models. Additional limitations of our approach that could explain the errors are: (1) the selection of features does not account for interactions between them, (2) the context of events is not taken into account: train speed, catenary tensions, etc. Nonetheless, the fact the models give similar and good performance for longer periods of time than the ones they were trained on, for three different datasets\footnote{which come from different vehicle manufacturers} gives confidence that the models generalize sufficiently well to be useful in practice.\\
 
Until now, train maintenance technicians need to investigate and to read lengthy event traces to explain a given incident. With the proposed approach, when a model gives the same conclusion as an expert, the explanation of incidents is captured by the sets of events extracted by the model. As a result, the confidence in the explanation is higher and and the extracted sets of events objectify it. These sets also give an idea of the recurring scenarios that lead to a delay, giving a leverage on potential actions to mitigate these.\\
When the model disagrees with train maintenance technician diagnostics, a special investigation is triggered by an expert to know what features where used by the train maintenance technician, whether he made a mistake or how the model could be improved. On the long term, this will allow to create a higher quality training data set to retrain the models. The adoption of the learning machine suggestions poses also a fundamental challenge  which is how do we make sure the train maintenance technicians keep both confidence in the learning machine and enough critical sense not to blindly trust the system ?

\section{Conclusions and future work}
The automation of railway vehicle incident diagnostics suggestions is proposed by formulating a classification task. The input data are events from traces generated on-board vehicles, which could be interpreted as the vocabulary of the language of railway vehicles. These are ingested and processed in a central cloud platform. A feature engineering approach is proposed to help experts extracting automatically sets of events that can explain the technical causes of incidents over fleets of vehicles rather than individually and manually reading sequences of events themselves. By sharing the predictions of the trained models in production to train maintenance technicians on their phones, tablets and laptops, they are able to have a better idea of what to repair on which vehicles before they even need to inspect them. A human feedback loop is implemented such that designated train maintenance experts can re-label data when the model and the train maintenance crews disagree. This feedback is used to re-train the models and maintain the performance on the long term.\\
Our approach accelerates and prioritizes the repair processes of vehicles which will increase the reliability of railway systems for potentially any train operator. Railway vehicle manufacturers could also benefit from the proposed approach during early design stages to define and improve which events need to be reported such that incidents can be expressed by vehicles.\\

Future work will explore how the extracted sets of events can be used to assist human experts in creating predictive maintenance alerts: not just to diagnose but to prevent incidents. In terms of modeling, the usage of LSTMs will be investigated. The performance of other classifiers will be evaluated. Finally, the long term relationship between a large crew of human workers and the learning machine will develop, needs to be investigated. 

\begin{acknowledgments}
 Funding comes from internal resources of SNCB-NMBS Engineering Department (B-TC.4).
\end{acknowledgments}

\bibliography{final_sncb_haii5}

\end{document}